\documentclass[conference]{IEEEtran}
\IEEEoverridecommandlockouts

\usepackage[margin=1in]{geometry} 
\usepackage{amsmath,amsthm,amssymb,hyperref, graphicx, tikz, afterpage}
\usepackage{mathtools,physics}
\usepackage{comment}
\usepackage{xcolor}
\usepackage{listings}
\usepackage{tikz}
\usetikzlibrary{positioning, arrows.meta, fit}

\tikzset{
  blk/.style  = {draw, rounded corners=3pt,
                 minimum height=1.05cm, minimum width=2.6cm,
                 font=\sffamily\small, align=center, fill=#1!10},
  blk/.default = gray!10,
  arr/.style   = {-{Stealth[length=3pt,width=3pt]}, very thin},
}

\usetikzlibrary{positioning, calc, arrows.meta}
\title{Machine-Learning Driven Load Shedding to Mitigate Instability Attacks in Power Grids}
\author{Justin Tackett$^{\dagger,1}$, Benjamin Francis, Luis Garcia$^\ddagger$, David Grimsman$^\dagger$, Sean Warnick$^\dagger$ \\\thanks{$^\dagger$Information and Decision Algorithms Laboratories (IDeA Labs), Brigham Young University, Provo, Utah $^\ddagger$University of Utah, Salt Lake City, Utah \textsuperscript{1} Corresponding author \texttt{jtackett@aht.ai, justin.carl.tackett@gmail.com}. All authors are also affiliated with Achilles Heel Technologies (AHT), and the work was  supported through AHT by the US Department of Energy, contract DE-SC0021693.  The views expressed here are entirely the authors' and do not necessarily reflect any official government position.}}

\begin{document}


\maketitle

\begin{abstract}
 Critical infrastructures are becoming increasingly complex as our society becomes increasingly dependent on them. This complexity opens the door to new possibilities for attacks and a need for new defense strategies. Our work focuses on instability attacks on the power grid, wherein an attacker causes cascading outages by introducing unstable dynamics into the system. When stress is place on the power grid, a standard mitigation approach is load-shedding: the system operator chooses a set of loads to shut off until the situation is resolved. While this technique is standard, there is no systematic approach to choosing which loads will stop an instability attack. This paper addresses this problem using a data-driven methodology for load shedding decisions. We show a proof of concept on the IEEE 14 Bus System using the Achilles Heel Technologies Power Grid Analyzer, and show through an implementation of modified Prony analysis (MPA) that MPA is a viable method for detecting instability attacks and triggering defense mechanisms.
\end{abstract}


\section{Introduction}
Throughout the past two hundred years, the power grid has become a core part of the infrastructure of the world. 
Every modern facility relies on electricity to sustain the way of life that has become prevalent in first world countries, powering everything from life sustaining equipment to financial transaction infrastructure. Recent events like the mass power outage in Spain \cite{lombardi2025granada}, widescale blackouts and brownouts every summer \cite{zeitlin2025summer}, and fears of future blackouts \cite{mitchell2025blackouts} serve as reminders that the power grid is delicate and without it, life as we know it comes apart at the seams.


With this potential disaster and its consequences in mind, it is a troubling idea that wide scale power outages could be purposefully brought about by terrorists or nation-state sponsored attackers, and a yet further troubling idea given that attacks like these have already happened at large scales \cite{industroyer, salazar2024tale}.

Because of this, it is critical to come to understand what kinds of attacks may be around the corner in the ever growing landscape of cyber-warfare targeting critical infrastructure, and how such attacks might be mitigated.

One sophisticated class of next generation cyber-attacks is instability attacks. These attacks connect potentially unrelated parts of a system and introduce feedback between them that destabilizes the entire system. Because it affects the entire system and can, in theory, be introduced anywhere in the system, the contact points can be extremely hard to detect\footnote{In fact, even the causes of non-malicious power outages can be extremely difficult to establish \cite{lombardi2025granada}.}, resulting in stubborn persistence with devastating impact \cite{sean1,sean2,sean3,sean4}.

 Because this class of attacks is so new, few, if any, strategies have been developed to mitigate them. Mitigation is particularly challenging given that the attack surface (the set of potentially exposed contact points) can be extensive and hard to cover. Currently, there are some efforts to try to raise awareness and provide solutions for these kind of attacks, but the general understanding of how to respond in the event of instability attacks is still in its early stages \cite{tackett}.

This paper presents a novel retrofitting of current remedial action mechanisms, namely load shedding mechanisms, using a cost-effective, data-driven approach to be able to respond to attacks before they do significant damage and interrupt them with a 0.92 overall F1-score. Methods for attack detection are also presented. These allow operators to train models on their own systems and establish software based logic to be able to respond to this new class of attack.

\section{Related Work }
The foundational paper for this work is Tackett et al. \cite{tackett}, which illustrates how instability attacks can be formed for a power grid, the control theory that supports their existence, and the effects it can have on a system. While similar works and case studies exist, \cite{destabmicro, DestabilizationStuxnet}, to our knowledge it is the first of its kind describing instability attacks on power systems by introducing modified or artificial feedback into the system to drive it to be small-signal unstable.

A much more extensive catalog exists regarding using load shedding with machine-learning, the approach taken in this paper. Schweitzer et al. \cite{Schweitzer}, in conjunction with their development of modified Prony analysis, use load shedding to combat unstable inter-area oscillations. This is seen in other works as well; Toro et al.~implemented automatic load shedding for the development of more robust remedial action schemes \cite{toro2020remedial}, which gives precedence for using load shedding without an human operator in the loop to combat growing oscillations. Justin and Paternain \cite{justin2024data} implement a reinforcement learning (RL) agent in conjunction with a classifier of system response given a certain input to do adaptive load shedding. This approach is very similar to the work presented here, and integrating the classifier here with a similar type of RL agent could prove useful in broad scale implementation with the hereafter mentioned 'operator-type' entity. The principal difference between their paper and this one is that we focus on training a classifier on instability attacks (with infrastructure to detect these attacks)---system responses that would almost certainly not be captured in traditional system configurations used in that paper. There are many other things happening in this space, a systematic survey of which was done recently by Skrjanc et al.\cite{skrjanc2023systematic}.

Cyber-physical anomaly and intrusion detection using machine learning is yet another application that has seen considerable work in recent years, both broadly for cyber-physical systems as well as specifically for power grids \cite{tushkanova2023detection,esmaeili2023anomaly,alani2023two,abshari2025survey}.


Modified Prony analysis (MPA) is a tool we have repurposed for objectives different than the original intent of inter-area oscillations \cite{Schweitzer}, however our application is conceptually close to theirs and so is presented as a secondary or minor contribution, with credit belonging to the Schweitzer Electronic Labs (SEL) for this tool. Others have also found and used this tool in similar areas to those used at SEL \cite{toro2020remedial}.

\begin{figure*}
  \centering
  \includegraphics[width=1.0\textwidth]{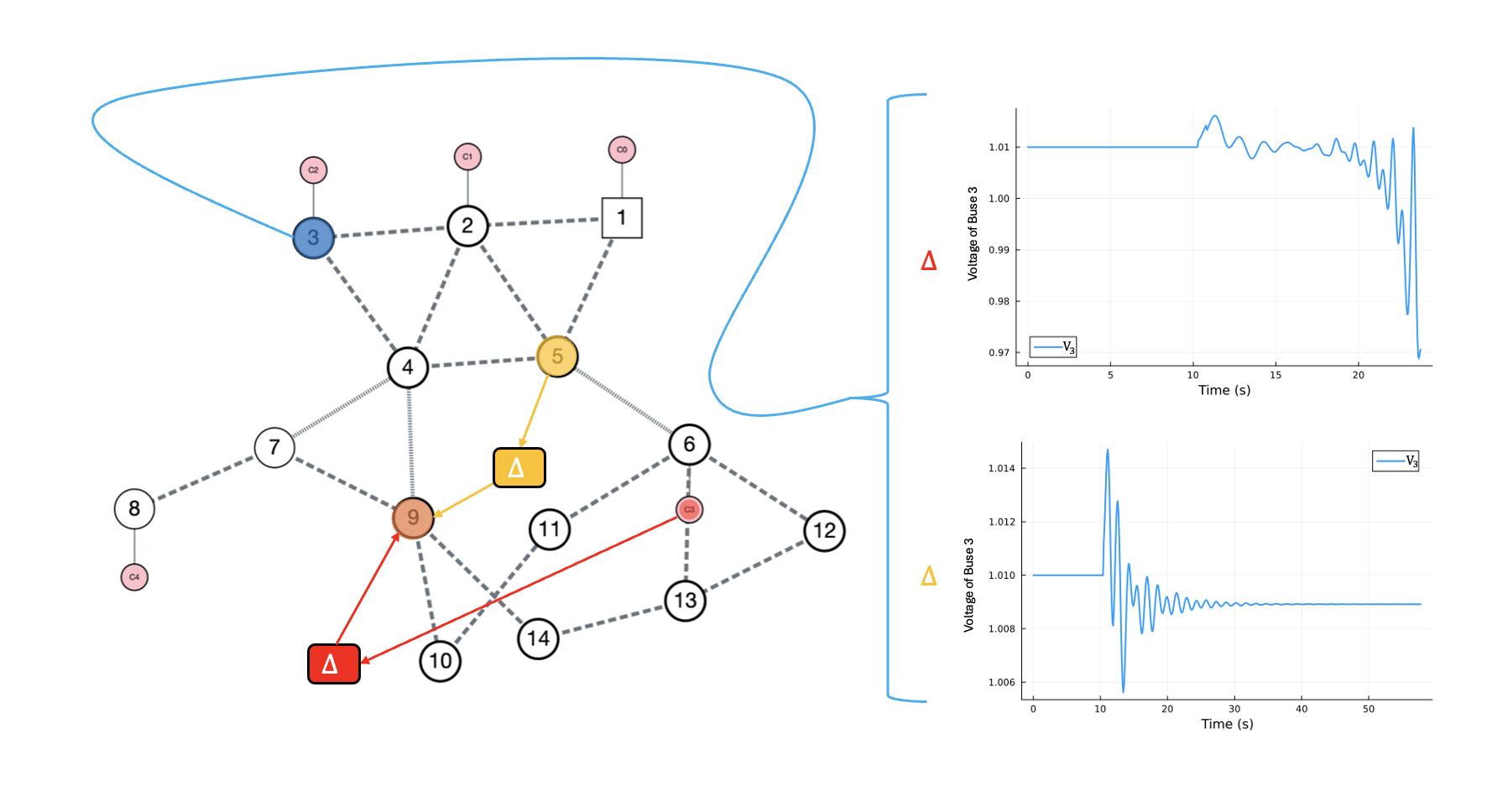}\\[1ex]
  \caption{An example of naively shedding according to under voltage load shedding schemes under an instability attack. On the left is the IEEE 14 Bus system, with generators as smaller red circles and buses as the white nodes. Two attacks are shown; a red $\Delta$ and a yellow $\Delta$, where each $\Delta$ reads from a state in a given node (Bus 5 and Generator 3 for yellow and red $\Delta$'s, respectively) and both write to the same state in Bus 9; these are all highlighted accordingly. These attacks happen separately, with the same response: shedding the load on Bus 6. The resulting effect on the system, shown by the voltage for Bus 3 (in normalized units), is shown in the two graphs to the right. In the instance of the red $\Delta$, the attack gets worse, and in the instance of the yellow $\Delta$, the attack is interrupted, thereby showing that using pre-defined load-shedding schemes can yield stable or unstable outcomes, and a system-aware decision process is needed.} 
  \label{fig:stacked}
\end{figure*}

\section{Methodology and Threat Model}
As a high-level anecdote to explain the tool, imagine that an adversary establishes the groundwork for an instability attack and launches it. Alarms are triggered by the MPA mechanisms that are normally used for operational grid stability but have had the relay logic adjusted slightly to detect these advanced attacks. Data begins to be captured and is fed into a machine-learning model (either cloud or edge based) that quickly provides an operator-type entity (be it an actual operator or another decision-making algorithm), a list of loads to shed that would stop the attack, and which loads (if shed) would worsen the attack. The entity then makes a decision, shedding a load, and the attack excites into a transient that passes, yielding way to normal post-load shedding procedures and steady-state stability.

This is one way the tool presented in this paper could be used and demonstrates the different parts at play here. We have an alarm system that uses MPA to detect an attack and trigger a machine-learning model. That model takes in system states for a given amount of time and yields a labeling for each load: stable shed, or unstable shed (as in, if we shed this load, will the system stabilize or get worse). That can then be used by an operator or a decision-making algorithm outside the scope of this paper to make a decision informed by all factors and stop an instability attack.

\subsection{Threat Model}
With this high level anecdote comes the question of threat modeling. This tool exists inside the context of defense in depth; for instability attacks to happen, a variety of other traditional cyber-securtity and physical safety mechanisms could potentially stop the attack. Instances where the attack is successfully stopped by these traditional methods are not those considered here. Instead, we consider situations where all other defenses have failed; perhaps a sufficiently sophisticated attacker is able to bypass other preventitive measures or uses previously unknown vulnerabilities to get to the point where they can launch an attack.

We therefore consider a zero-trust environment, where we assume that at least two nodes in our system have been compromised and we do not know which ones. Our action is to shed loads (through physical actions like flipping breakers), and we assume our capacity to do so has largely not been compromised. This assumption is somewhat fair, as the attacker likely must use the same communication network and other infrastructure to carry out the attack that a defender would use to stop the attack, and so it likely cannot be completely disabled, and even if it is, we consider load shedding opportunities across the power grid, and compromising them at scale is a scenario not within the scope of this threat model.

\subsection{Motivating Example}

As a motivating example, consider the IEEE 14 Bus System shown in Figure \ref{fig:stacked}, a popular research benchmark in the power systems community for studying a variety of phenomena, including stability. Physically, shedding a load is a matter of a breaker being tripped, ceasing the draw of power for a given community on the grid. In our simulation of the IEEE 14 Bus System (the dynamics for which can be found in
Appendix \ref{equations}),
this amounts to setting the load states (the active and reactive loads, referred to in this paper as $PL$ and $QL$) on a given bus to zero at a discrete point in time. In traditional load shedding schemes, loads to be shed are chosen a priori to any event, and are almost entirely dependent on the topology of the system. If we take a similar approach and shed the same bus in the same topology with the same initial conditions, the traditional approach suggests that we would get the same outcome. But the reality, as seen in Figure \ref{fig:stacked}, is that for two different instability attacks, shedding the same bus gives different results: one outcome where the system returns to stability, and the other where the attack is amplified and even exacerbated. This shows that choosing a load to shed under an instability attack is a non-trivial problem, and needs to be addressed. One way to address this is through a data-driven approach.

\subsection{Data Source and Processing}

For a data-driven approach, we need data. To show a proof-of-concept tool that demonstrates the power and efficacy of data-driven loadshedding to bypass instability attacks, we used the Achilles Heel Technologies (AHT) Power Grid Analyzer, a tool that can take in a model of power grid and analyze its vulnerability to instability attacks, including modeling the results of these attacks. We modified this tool to implement load shedding by cutting to zero at a discrete time in the numerical solver the values for a given load. While no model is a substitute for reality, this numerical simulation of load shedding under an instability attack yields valuable insight and promise for further developed tools to be based on this approach. We used the generic, predefined generator models and parameter values given in the software tool to model the IEEE 14 Bus System. More about the tool can be found in Tackett et al. \cite{tackett}.

For the sake of training our machine-learning model, there are essentially three pieces of information we start off with: 1) which load was shed, 2) the system states for the attack (before and after load shedding) and 3) the states being read or written to by $\Delta$. The data we would eventually want to feed into the model for training then consisted of three somewhat separate parts: 1) which load would be shed, 2) the system states before the load shed, and 3) whether or not the load shed would lead to a stable outcome. We obtained the first two by iterating through each type of attack and every load shedding possibility, simulating each such scenario, and truncating the data at a well-defined pre-load shedding point.

\subsubsection*{Modified Prony Analysis (MPA)}

Next, we considered how the process to trigger the model. A method used for finding unstable inter-area oscillations proved to be able to establish an ‘alarm’ of sorts to alert a system that an attack is underway. 
This method is called the modified Prony analysis (MPA), presented by the Schweitzer Engineering Lab (SEL) \cite{Schweitzer}. 
The paper presents a method for a data-driven method (not machine-learning) that establishes whether or not a given signal has an unstable mode. 
Prony analysis was originally used in ringdown analysis, where an impulse would be put into a system, and the modes would then be analyzed.  It does this by doing a parametric curve fitting that leverages that any signal can be approximated by damped complex exponentials. 
Modified Prony analysis does a similar thing, but instead of using an impulse to analyze stable modes, it looks at a sliding window of the data and identifies unstable modes early on. It does this in an iterative way that rejects noise modes according to some tolerances and hyperparameters, as described in the SEL paper \cite{Schweitzer}.

Using MPA, we can determine whether an instability attack has created an unstable mode. We coded up and implementation, and saw success in detecting the instability attacks that we provided it, and this code is included with the ML model code. MPA is used already in power grid settings for unstable interarea oscillation detection, and so modifying the existing infrastructure slightly would allow instability attack detection and integrate into the tool we are developing as the alarm signal that would start the analysis.


In future work and for real systems, MPA is a strong candidate for labeling training data, however, for this specific case, a heuristic labeling system was used. The heuristic labeling system, described in the following section, would likely fail on noisy data and so is not as competitive an alternative to MPA for labeling training data.

\subsubsection*{Heuristic Label}

The heuristic label\footnote{The heuristic label used here is specifically designed for the quirks and limitations of the numerical integrator and edge case handling of the AHT Power Grid Analyzer and are presented for clarity alone. 
If a reader would like all the code used for the generation of data and the heuristic that is then applied to get the training labels, such code is available upon request to the author for reproducibility’s sake, with exceptions and corresponding concessions applying to where the AHT Power Grid Analyzer source code is proprietary.} can be divided up into three tests: 
\begin{enumerate}
    \item Absolute-excursion test: This test checks to see if the scaled values of a given state go below or above a certain threshold; if they do (e.g. a voltage dip below 80\% of its optimal value) then it is almost certain that instability of some kind will follow, normally with some form of cascading failure. This almost always (if not 100\% of the time) corresponds to some major failure of the load shed, as it means the instability was magnified significantly to reach those values. In these cases, the test returns back that it was an unstable shed at this point.
    \item Variance Test: Looking at the last 25\% of the signal, if there is little to no variance, this almost always is indicative of some kind of stability, even if the new equilibrium is different from the original one slightly. If this is the case, at this point the heuristic returns true, because if not the next test may incorrectly label it as unstable
    \item Envelope-slope test: The last 25\% of the signal is split into six equal windows. The RMS magnitude is computed in each window, and a least squares line is fitted to the log of the RMS across the six windows. If the slope isn’t below a certain negative threshold, it is deemed unstable.
\end{enumerate}
With an input of the post-load shed data, this labeler was used to label the pre-load shed data with the corresponding load shed index as stable or unstable.

\subsubsection*{Data Processing}
\label{sec:dataprocessing}
With the heuristic label in hand and now having all the tools to generate the data we want, we created a multi-threaded script to process all of the data and put it into training, testing, and validation data sets for the machine-learning model.

It is worth noting that because of bugs with the numerical integrator, the AHT Power Grid Analyzer, or other unknown sources of bugs, not every single configuration of load and attack type yielded viable data. So as a way to remove low quality data, we made a few decisions.

First, if the integrator stopped after the load shedding for any reason (numerical stiffness being a principal culprit), that wouldn’t work because there would be no way to label the data. Further to this aspect, if the data didn’t go on for at least 50 seconds\footnote{Preliminary surveying the parameter space found that at stages of the attack when the signal was still small, the response to load shedding was independent of time, so ten seconds after the attack was launched was used as the universal load shedding time for data generation and model training.} after the load shed (most would go to about 200), then we decided not to use it as labeling would likely be inconclusive. Finally, if the integrator just ran into problems for some reason, we decided to see how much data we could successfully get and if it was sufficient to not need to address the bugs that caused those to fail.

\begin{figure*}
    \centering
    \includegraphics[width=1.0\textwidth]{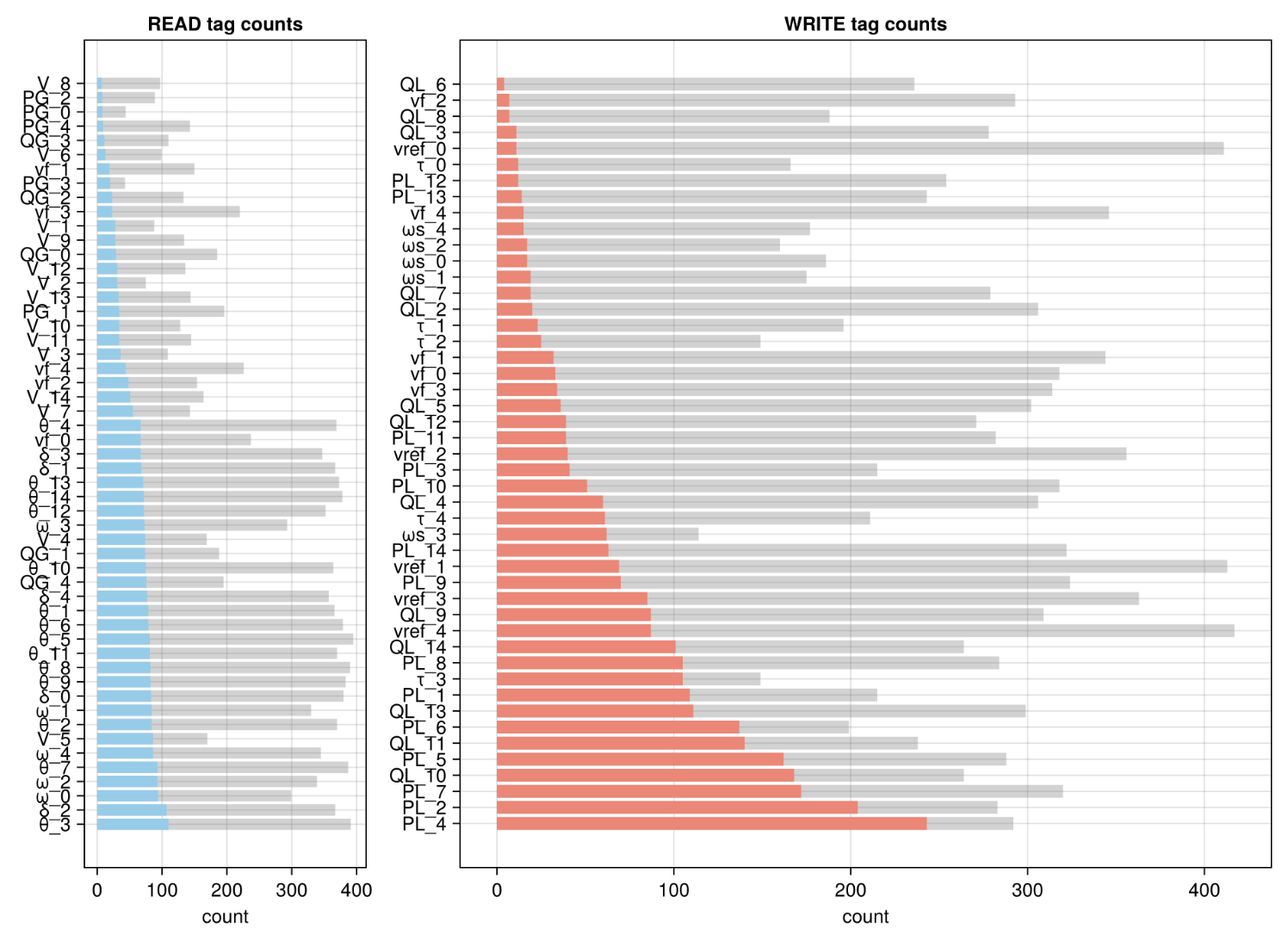}
    \caption{A comparison of the distributions of stable shed cases across different variable categories. These are taken from the 12,808 viable data samples used for training, testing and validation. They are shown to show that each read/ write combination is represented in the stable case training sets, and also showing the relative number of cases present. For a given bar, say reading Voltage for bus 3 (V 3), all possible write combinations and all load shedding options were summed together to a bar, where the colored bar is the count of stable load sheds and the grey bar is the total amount of samples for the given category, providing some insight into sample distribution. This is likewise for the right concerning write variables. The labels here represent different physical variables such as frequency $\omega$, rotor angle $\delta$, voltage $V$, load angle $\theta$, reactive and active loads and power generation $PL,QL,PG,QG$, and so on for generators/ loads $i$ for the IEEE 14 Bus System.}
    \label{fig:stable_dist}
\end{figure*}

Ultimately, we ended up with about 12,808 viable samples, which was determined to enough to be split among the training, testing, and validation datasets. While this sample size was large enough for our purposes of creating a proof of concept model, in future work, we're looking to address the numerical integrator to get a more comprehensive sampling of the entire attack configuration space, however as evidenced in Figure \ref{fig:stable_dist}, we were able to capture a decent sampling of every category of attack.

\section{Model Construction and Training}
As alluded to earlier, we determined a classifier would be the best choice for the tool. The data (load index and ten seconds of pre-shed system states) would be fed into the model, which would yield a boolean label: stable or unstable shed. 

We did this by making a time-series encoder and a load index encoder, which then are both fed together into a fully connected (FC) three layer MLP of dimensions 128, 64, and 2, respectively. This yields soft-max logits, or probabilities, of each classification. When doing inference, these values are then compared against some confidence threshold $\tau$ to determine which classification will be made. By lowering $\tau$, more riskier stable sheds are just marked as unstable, making the system safer overall as false positive stable shed classifications are much more dangerous than false positive unstable sheds.

The two encoders are fairly straightforward; in the case of the time-series encoder, the time series with each channel (voltage, frequency, etc.) are combined into one $C \times T$ matrix. One dimensional convolution is done on each channel twice to be able to capture features and allow sufficient expressivity. These are then put through a MaxPool, taking a two-sample window across the sample, reducing noise and reducing the computational load for the Bi-GRU. Then these are fed into the Bi-GRU which, as the name implies, runs through the sample forward and backward, and keeps $H$ dimensions for each run, concatenated together for $h_{\mathrm{ts}}$ of size $2 \times H$, the output of the time-series encoder.

The load-index encoder is even more straightforward, having just a one layer MLP that takes the load index value and puts it into 32 layers, allowing for more complex, non-linear features in the load's relationship with whether a stable or unstable classification is given. The result of this one-layer MLP $h_{\mathrm{ld}}$ is then concatenated with $h_{\mathrm{ts}}$ to be fed into the FC three-layer MLP. The result of these are fed into a soft-max (during training, they're fed into a cross-entropy loss which internally applies the soft-max).

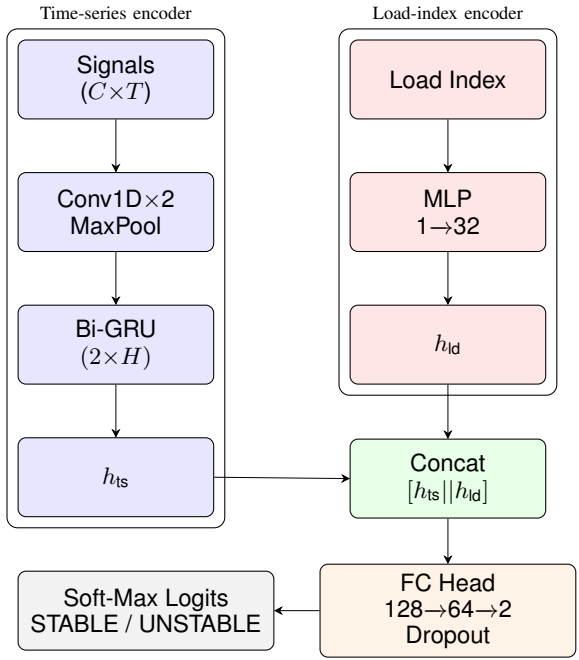
\begin{figure}[t]
  \centering
  \begin{tikzpicture}[node distance = 7mm and 18mm]

    \node[blk=blue]   (sig)  {Signals\\($C{\times}T$)};
    \node[blk=blue,below=of sig] (conv) {Conv1D$\times$2\\MaxPool};
    \node[blk=blue,below=of conv] (gru)  {Bi-GRU\\$(2{\times}H)$};
    \node[blk=blue,below=of gru]  (hts)  {$h_{\text{ts}}$};

    \foreach \a/\b in {sig/conv, conv/gru, gru/hts}
        \draw[arr] (\a) -- (\b);

    \node[blk=red,right=of sig]  (load) {Load Index};
    \node[blk=red,below=of load] (mlp)  {MLP\\1$\xrightarrow{}$32};
    \node[blk=red,below=of mlp]  (hld)  {$h_{\text{ld}}$};

    \foreach \a/\b in {load/mlp, mlp/hld}
        \draw[arr] (\a) -- (\b);

    \node[blk=green,
          below=12.5mm of hld -| hld.center] (cat)
          {Concat\\$[h_{\text{ts}}||h_{\text{ld}}]$};

    \node[blk=orange,
          below=6mm of cat,
          minimum width=3.4cm] (head)
          {FC Head\\128$\xrightarrow{}$64$\xrightarrow{}$2\\Dropout};

    \node[blk=gray,
          left=6mm of head,
          minimum width=3.4cm] (out)
          {Soft-Max Logits\\STABLE / UNSTABLE};

    \draw[arr] (hts.east) -- (cat.west);
    \draw[arr] (hld) -- (cat.north);
    \draw[arr] (cat) -- (head);
    \draw[arr] (head) -- (out);

    \node[draw,rounded corners=4pt,inner sep=4pt,
          fit=(sig)(conv)(gru)(hts),
          label={[font=\scriptsize]above:Time-series encoder}] {};

    \node[draw,rounded corners=4pt,inner sep=4pt,
          fit=(load)(mlp)(hld),
          label={[font=\scriptsize]above:Load-index encoder}] {};

  \end{tikzpicture}

  \caption{The block diagram for the classifier used to determine if a load shed results in stable conditions or not. The classifier is made up of principally three parts: a time-series encoder, a load-index encoder, and a fully connected three layer MLP for classification. The output is passed through soft-max logits, which provide probabilities of classification as stable or unstable. For the IEEE 14 Bus System, this results in about 600,000 weights, relatively light-weight and able to be trained on a laptop in several hours.}
  \label{fig:model}
\end{figure}

\section{Results }
\label{sec:results}
\subsection{Model Results}
After training the model for several hours on a Macbook Pro with an M3 Pro chip and 18 gigabytes of memory, the model was able to correctly classify loads with an overall F1-score of 0.92, as shown in detail in Table \ref{tab:accuracy}.

\begin{table}[ht]
  \centering
  \caption{Classification Report, $\tau$ = 0.03}
  \label{tab:accuracy}
  \begin{tabular}{lcccc}
    \hline
    & Precision & Recall & F1-score & Number of Cases \\
    \hline
    STABLE       & 0.95 & 0.68 & 0.79 &  599 \\
    UNSTABLE     & 0.91 & 0.99 & 0.95 & 1962 \\
    \hline
    Accuracy     & & & 0.92 & 2561 \\
    Macro avg    & 0.93 & 0.84 & 0.87 & 2561 \\
    Weighted avg & 0.92 & 0.92 & 0.91 & 2561 \\
    \hline
  \end{tabular}
\end{table}

Since we are dealing with critical systems, minimizing false positives for stable labels is crucial; having slightly less options for stable load shed is much better than making a decision thinking it. This corresponds to maximizing stable precision; for this table, precision means of all the cases we identified with this label, how many were correctly identified, where recall means of all the cases that were labeled, how many did we get.

A hyperparameter $\tau$ exists that allows us to essentially draw a line of how conservative we would like to be in our labeling. A lower $\tau$ value means that we throw away more potentially stable cases in order to keep stable precision high, and so $\tau$ = 0.03 brings us up to a precision of 95\% while keeping 68\% of the total stable cases. Given that of all the load shedding cases, 23\% of the cases result in stable sheds, this moves us to approximately 16\%, meaning that in large systems, there will more than likely be at least a handful of viable load shedding options. The successful load sheds largely resemble that seen in Figure \ref{fig:stacked}(b).

\subsection{Data Statistics}
Collecting some metrics from the base cases, we can see some interesting trends (or lack thereof) further confirming a data-driven solution as the appropriate approach.

\begin{figure}[h]
    \centering
    \includegraphics[width=0.5\textwidth]{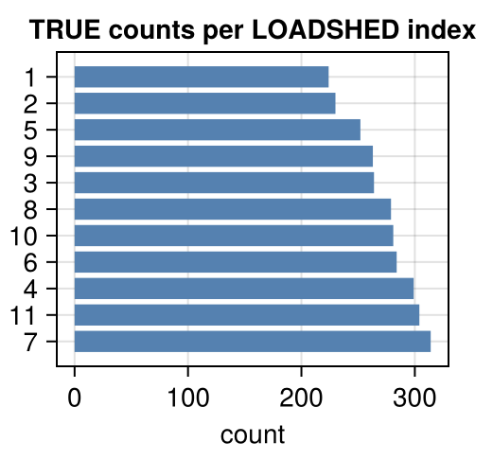}
    \caption{A distribution of stable sheds for all possible read/ write attack configurations, per load. For example, load 2 here represents how many times in the 12,808 samples that picking load 2 resulted in stability. This graph shows that, barring some small variance, there isn't one load that picking every time would result in stable outcomes, showing the need for a data-driven approach over a priori load shedding schemes. }
    \label{fig:loadshedindx}
\end{figure}

First, we can see in Figure \ref{fig:loadshedindx} that across all 12,808 cases, choosing the correct load to shed was much more than purely topology dependent, standing in stark contrast to traditional load shedding schemes that are almost entirely network topology dependent. The cause for this is likely the mathematical differences behind traditional load shedding and load shedding to defend against instability attacks; in the former case, you are doing something like rebalancing the swing equation, while in the latter, you are modifying the network structure to disrupt unstable modes.

Next, we can see that while we could only consider about half of the total attack/ load configurations, every read and write variable was able to be represented in some way. It is probable that those lower in representation were affected by issues that caused many of them to be cut, like mentioned in Section \ref{sec:dataprocessing}, and this could also affect the cases where there is a significantly lower amount of successful load sheds present, but also may be indicative of some attacks just being harder to defend against with single load shedding than others. This is a topic of great interest, and will be explored in future work.

It is worth mentioning that while we were able to, at scale, get good representation of each category and these seem to suggest at least one load being able to be shed for any given attack, this is not the case. There are many attacks for the IEEE 14 Bus System where no load shed is effective, and others where most load sheds are effective. The average case is that at least one or two work are effective, but this difference further highlights some of the interesting aspects mentioned earlier about some attacks perhaps being harder to defend against than others, and will be explored more in future work.

\section{Discussion }
The primary contributions of this paper are the model infrastructure in code and the proof of concept of this model on the IEEE 14 Bus System using the AHT Power Grid Analyzer.
A secondary contribution that serves more as an extension of research done by others is the identification of the MPA as an effective alarm system and data labeler to train and trigger a classifier on early stage attack data.
These tools do some things well, while leaving other areas for future research.
\subsection{Strengths}
The model does well in a couple regards that we had mentioned before: by using a data-driven approach, we sidestep computational time requirements, model variance issues (inasmuch as the system trained on is allowed to vary), and complexity issues by utilizing supervised machine-learning to create a labeler based on system states. 

Instead of identifying only one load to shed, the labeler instead identifies every viable load in a system, allowing an operator-type entity to weigh the pros and cons of other considerations with load shedding, like load importance to customers, impact on system recovery post attack, and types of stability outside of small signal stability that instability attacks target.

Finally, the implementation of this approach is relatively inexpensive. Unlike other approaches for mitigating against instability attacks, such as restructuring power grid topology, which require physical restructuring of the system, this approach is a software based tool that repurposes existing safety and contingency equipment to be able to defend against instability attacks, lowering overhead costs comparatively.

\subsection{Weaknesses}
Because this is a data-driven approach, the principle weakness for this approach surrounds the availability, latency, and integrity of data.

Cloud connected industrial control system hardware is becoming more commonplace, and this certainly helps in addressing this problem, but the approach as it currently stands needs system measurements for the entire system, and so if there are legacy components, this proves an issue as data collection systems (e.g. SCADA) typically don't have the temporal resolution necessary to respond to instability attacks because of the differing timescales of data collection and attack dynamics.
Low level controllers and relays collect data at a sufficiently fine scale to be able to detect attacks and could trigger a higher resolution tracking in the data collection system if MPA detects an attack, and cloud infrastructure provides a possible solution, but both of these are still subject to the issue of latency.

Because power grids are often times geographically spread out, the latency between events and those events reaching SCADA typically means that emergency responses need to happen locally and then reported back to the central system.
Naturally this leads to the question, can this kind of model be employed as a series of decentralized agents?
While an interesting subject for future work, one concern is that if all or many agents separately decided to shed loads separately, this could create wide scale grid instability, which is the principal reason that the classifier was trained as if only one load could be shed: losing one load is equivalent to a normal power outage, losing many loads brings up issues of grid stability.
These problems fall into the popular area of decentralized multi-agent systems, and applying the results from that field to these results is another potential area to go with this work.

Finally, there is the issue of data integrity.
This approach is only as good as the integrity of the data coming back, and so if whatever process connects sensors to the alarm system or model is spoofed so as to not give an accurate picture of system states, then this mitigation attempt fails.
Wide scale sensor spoofing is unlikely and impractical for large power grids, and so the main weakness here is how spoofing one sensor affects the classifiers ability to correctly identify the load to shed.
This was not considered in this paper and is a topic for future work.
One potential solution would be to allow adversarial inputs into the model to increase robustness, but this would inevitably come at the cost of performance as it does with any classifier \cite{robustperformancetradeoff}, and therefore is a fundamental weakness of a data-driven approach.

\section{Conclusion}
While instability attacks are not prevalent today, in a world of critical infrastructure increasingly becoming a target in cyberwarfare, their discovery and implementation by powers that wish to do harm seems inevitable, so finding ways to mitigate against them is critical. We have presented a data-driven model that can use load shedding systems in a power grid to identify how to respond against these attacks, laying down the foundation and proving the concept for industry level tools to apply these approaches at scale and identify ways to prevent these attacks from doing serious damage to critical infrastructure around the world. We have shown that naive load shedding can be disastrous, and given tools for detecting these attacks as well as labeling their effects post load shedding. We have identified strengths and weaknesses, and identified ways these systems must be improved for the practical problem of implementing the solution at scale.
\begin{appendices}
\section{Power Grid Dynamic System}
\label{equations}
Generator equations (with AVR):
\begin{align}
       v_d &= V\sin(\delta - \theta) \\
       v_q &= V\cos(\delta - \theta) \\
       P_g &= v_d i_d + v_q i_q \\
       Q_g &= v_q i_d - v_d i_q \\
    \tau_e &= (r_a i_q + v_q)i_q + (r_a i_d + v_d)i_d \\
    \dot{\delta} &= \Omega_b(\omega - \omega_s) \\
	\dot{\omega} &= \frac{1}{2H}(\tau_m - \tau_e - D(\omega - \omega_s)) \\
    \dot{e}'_q  &= \frac{1}{T'_{d0}}(-e'_q - (x_d - x'_d)i_d + v_f) \\
	\dot{e}'_d  &= \frac{1}{T'_{q0}}(-e'_d + (x_q - x'_q)i_q) \\
    0 &= v_q + r_a i_q - e'_q + x'_d i_d \\
	0 &= v_d + r_a i_d - e'_d - x'_q i_q
\end{align}
\begin{align}
    \dot{v}_m    &= \frac{1}{T_m}\left(V - v_m\right) \\
    \dot{v}_{r1} &= \frac{1}{T_a}\left(K_a\left(v_\text{ref} - v_m - v_{r2} - \frac{K_f}{T_f}v_f\right) - v_{r1}\right) \\
    \dot{v}_{r2} &= -\frac{1}{T_f}\left(\frac{K_f}{T_f}v_f + v_{r2}\right) \\
    \dot{v}_f    &= -\frac{1}{T_e}\left(v_f\left(K_e + A_e e^{B_e\abs{v_f}}\right) - v_r\right) \\
    v_r &= \begin{cases}
                v_r^\text{min} & v_{r1} < v_r^\text{min} \\
                v_{r1} & v_r^\text{min} < v_{r1} < v_r^\text{max} \\
                v_r^\text{max} & v_{r1} > v_r^\text{max}.
            \end{cases}
\end{align}
Bus equations ($i$ is bus index):
\begin{align}
	0 &= P_{g,i} - P_{\ell,i} + \sum_k P_{b,ik} \\
	0 &= Q_{g,i} - Q_{\ell,i} + \sum_k Q_{b,ik}.
\end{align}
Transmission line equations:
\begin{align}
    P_{b,ik} &= Y_{ik} V_i (V_k\cos(\phi_{ik} + \theta_i - \theta_k) - V_i\cos(\phi_{ik})) \\
	Q_{b,ik} &= Y_{ik} V_i (V_k\sin(\phi_{ik} + \theta_i - \theta_k) - V_i\sin(\phi_{ik})) + b_{ik}V_i^2.
\end{align}
Transformer sending side equations:
\begin{align}
	P_{b,ik} &= Y_{ik} \frac{V_i}{m_{ik}} \left(V_k\cos(\phi_{ik} + \theta_i - \theta_k) - \frac{V_i}{m_{ik}}\cos(\phi_{ik})\right) \\
	Q_{b,ik} &= Y_{ik} \frac{V_i}{m_{ik}} \left(V_k\sin(\phi_{ik} + \theta_i - \theta_k) - \frac{V_i}{m_{ik}}\sin(\phi_{ik})\right) + b_{ik}V_i^2.
\end{align}
Transformer receiving side equations:
\begin{align}
	P_{b,ik} &= Y_{ik} V_i \left(\frac{V_k}{m_{ik}}\cos(\phi_{ik} + \theta_i - \theta_k) - V_i\cos(\phi_{ik})\right) \\
	Q_{b,ik} &= Y_{ik} V_i \left(\frac{V_k}{m_{ik}}\sin(\phi_{ik} + \theta_i - \theta_k) - V_i\sin(\phi_{ik})\right).
\end{align}

\end{appendices}
\bibliographystyle{IEEEtran}
\bibliography{ref.bib}

\begin{thebibliography}{10}
\providecommand{\url}[1]{#1}
\csname url@samestyle\endcsname
\providecommand{\newblock}{\relax}
\providecommand{\bibinfo}[2]{#2}
\providecommand{\BIBentrySTDinterwordspacing}{\spaceskip=0pt\relax}
\providecommand{\BIBentryALTinterwordstretchfactor}{4}
\providecommand{\BIBentryALTinterwordspacing}{\spaceskip=\fontdimen2\font plus
\BIBentryALTinterwordstretchfactor\fontdimen3\font minus \fontdimen4\font\relax}
\providecommand{\BIBforeignlanguage}[2]{{%
\expandafter\ifx\csname l@#1\endcsname\relax
\typeout{** WARNING: IEEEtran.bst: No hyphenation pattern has been}%
\typeout{** loaded for the language `#1'. Using the pattern for}%
\typeout{** the default language instead.}%
\else
\language=\csname l@#1\endcsname
\fi
#2}}
\providecommand{\BIBdecl}{\relax}
\BIBdecl

\bibitem{lombardi2025granada}
\BIBentryALTinterwordspacing
P.~Lombardi. Granada substation power loss pinpointed as ground zero of spain's blackout. Reuters. Accessed 2025-07-10. [Online]. Available: \url{https://tinyurl.com/472nbvms}
\BIBentrySTDinterwordspacing

\bibitem{zeitlin2025summer}
\BIBentryALTinterwordspacing
M.~Zeitlin. Blackouts, brownouts, and freaked-out grid operators: The summer of load has arrived. Heatmap News. Accessed 2025-07-10. [Online]. Available: \url{https://heatmap.news/energy/summer-of-load}
\BIBentrySTDinterwordspacing

\bibitem{mitchell2025blackouts}
\BIBentryALTinterwordspacing
A.~Mitchell. Britain preparing for blackouts on the scale of power cuts seen in spain and portugal, resilience plan reveals. The Independent. Accessed 2025-07-10. [Online]. Available: \url{https://www.independent.co.uk/news/uk/politics/uk-power-cuts-spain-portugal-blackouts-resilience-plan-b2784940.html}
\BIBentrySTDinterwordspacing

\bibitem{industroyer}
P.~Kozak, I.~Klaban, and T.~Šlajs, ``Industroyer cyber-attacks on {Ukraine's} critical infrastructure,'' in \emph{2023 International Conference on Military Technologies (ICMT)}, 2023, pp. 1--6.

\bibitem{salazar2024tale}
L.~Salazar, S.~R. Castro, J.~Lozano, K.~Koneru, E.~Zambon, B.~Huang, R.~Baldick, M.~Krotofil, A.~Rojas, and A.~A. Cardenas, ``A tale of two industroyers: It was the season of darkness,'' in \emph{2024 IEEE Symposium on Security and Privacy (SP)}.\hskip 1em plus 0.5em minus 0.4em\relax IEEE, 2024, pp. 312--330.

\bibitem{sean1}
A.~Rai, D.~Ward, S.~Roy, and S.~Warnick, ``Vulnerable links and secure architectures in the stabilization of networks of controlled dynamical systems,'' in \emph{2012 American Control Conference (ACC)}.\hskip 1em plus 0.5em minus 0.4em\relax IEEE, 2012, pp. 1248--1253.

\bibitem{sean2}
V.~Chetty, N.~Woodbury, E.~Vaziripour, and S.~Warnick, ``Vulnerability analysis for distributed and coordinated destabilization attacks,'' in \emph{53rd IEEE Conference on Decision and Control}.\hskip 1em plus 0.5em minus 0.4em\relax IEEE, 2014, pp. 511--516.

\bibitem{sean3}
V.~Chetty, S.~Warnick, S.~Roy, and S.~Das, ``Meanings and applications of structure in networks of dynamic systems,'' in \emph{Principles of cyber-physical systems: an interdisciplinary approach}.\hskip 1em plus 0.5em minus 0.4em\relax Cambridge Univ. Press, 2020, pp. 162--201.

\bibitem{sean4}
D.~Grimsman, V.~Chetty, N.~Woodbury, E.~Vaziripour, S.~Roy, D.~Zappala, and S.~Warnick, ``A case study of a systematic attack design method for critical infrastructure cyber-physical systems,'' in \emph{2016 American Control Conference (ACC)}.\hskip 1em plus 0.5em minus 0.4em\relax IEEE, 2016, pp. 296--301.

\bibitem{tackett}
J.~Tackett, B.~Brown, B.~L. Francis, T.~N. Burrows, M.~K. Transtrum, D.~Grimsman, and S.~Warnick, ``Automated red-teaming for securing critical infrastructures against instability attacks,'' in \emph{2024 IEEE Power \& Energy Society General Meeting (PESGM)}, 2024, pp. 1--5.

\bibitem{destabmicro}
J.~Zhou, X.~Chen, L.~Yan, and J.~Wen, ``Resilient distributed voltage control against destabilization attacks in {DC} microgrids,'' in \emph{2021 IEEE 5th Conference on Energy Internet and Energy System Integration (EI2)}, 2021, pp. 1163--1168.

\bibitem{DestabilizationStuxnet}
\BIBentryALTinterwordspacing
H.~Sandberg, V.~Gupta, and K.~H. Johansson, ``Secure networked control systems,'' \emph{Annual Review of Control, Robotics, and Autonomous Systems}, vol.~5, no.~1, pp. 445--464, 2022. [Online]. Available: \url{https://doi.org/10.1146/annurev-control-072921-075953}
\BIBentrySTDinterwordspacing

\bibitem{Schweitzer}
Y.~Gong and A.~Guzman, ``Synchrophasor-based online modal analysis to mitigate power system interarea oscillation,'' in \emph{2009 Distributech Conference Proceedings}, 2009.

\bibitem{toro2020remedial}
M.~Toro, J.~Segundo, C.~Nu{\~n}ez, N.~Visairo, and A.~Esparza, ``Remedial action scheme based on automatic load shedding for power oscillation damping,'' in \emph{2020 IEEE International Autumn Meeting on Power, Electronics and Computing (ROPEC)}, vol.~4.\hskip 1em plus 0.5em minus 0.4em\relax IEEE, 2020, pp. 1--6.

\bibitem{justin2024data}
G.~Justin and S.~Paternain, ``Data-driven under frequency load shedding using reinforcement learning,'' \emph{arXiv preprint arXiv:2410.04316}, 2024.

\bibitem{skrjanc2023systematic}
T.~Skrjanc, R.~Mihalic, and U.~Rudez, ``A systematic literature review on under-frequency load shedding protection using clustering methods,'' \emph{Renewable and Sustainable Energy Reviews}, vol. 180, p. 113294, 2023.

\bibitem{tushkanova2023detection}
O.~Tushkanova, D.~Levshun, A.~Branitskiy, E.~Fedorchenko, E.~Novikova, and I.~Kotenko, ``Detection of cyberattacks and anomalies in cyber-physical systems: Approaches, data sources, evaluation,'' \emph{Algorithms}, vol.~16, no.~2, p.~85, 2023.

\bibitem{esmaeili2023anomaly}
F.~Esmaeili, E.~Cassie, H.~P.~T. Nguyen, N.~O. Plank, C.~P. Unsworth, and A.~Wang, ``Anomaly detection for sensor signals utilizing deep learning autoencoder-based neural networks,'' \emph{bioengineering}, vol.~10, no.~4, p. 405, 2023.

\bibitem{alani2023two}
M.~M. Alani, L.~Mauri, and E.~Damiani, ``A two-stage cyber attack detection and classification system for smart grids,'' \emph{Internet of Things}, vol.~24, p. 100926, 2023.

\bibitem{abshari2025survey}
D.~Abshari and M.~Sridhar, ``A survey of anomaly detection in cyber-physical systems,'' \emph{arXiv preprint arXiv:2502.13256}, 2025.

\bibitem{robustperformancetradeoff}
D.~Tsipras, S.~Santurkar, L.~Engstrom, A.~Turner, and A.~Madry, ``Robustness may be at odds with accuracy,'' \emph{arXiv preprint arXiv:1805.12152}, 2018.

\end{thebibliography}

\end{document}